\documentclass{article}

\usepackage{arxiv}

\usepackage[utf8]{inputenc} 
\usepackage[T1]{fontenc}    
\usepackage{hyperref}       
\usepackage{url}            
\usepackage{booktabs}       
\usepackage{amsfonts}       
\usepackage{nicefrac}       
\usepackage{microtype}      
\usepackage{lipsum}		
\usepackage{graphicx}
\usepackage{natbib}
\usepackage{doi}
\usepackage{orcidlink}
\usepackage{authblk}
\usepackage{xspace}
\usepackage{enumitem}
\usepackage{amsmath}
\usepackage{multirow}
\usepackage{subcaption}
\usepackage[normalem]{ulem}
\useunder{\uline}{\ul}{}

\newcommand*{\fullnameSeisB}{All-Block Distillation\xspace} 
\newcommand*{\nameSeisB}{ABD\xspace} 

\newcommand*{\fullnameUnoB}{End-Block Distillation\xspace} 
\newcommand*{\nameUnoB}{EBD\xspace} 
\xspaceaddexceptions{\_}

\title{Too much of a good thing -- when knowledge distillation promotes overfitting, and how to avoid it}

\author{
    \normalsize\bfseries
    Irene Trigueros-Lorca\textsuperscript{2}\thanks{Corresponding author: \texttt{irenetrigueros@ugr.es}}\;\,\orcidlink{0009-0001-8836-2481} \quad
    Leonardo Concepción\textsuperscript{2}\,\orcidlink{0000-0002-4515-2038} \quad
    Christian Wagner\textsuperscript{3}\,\orcidlink{0000-0002-6121-9722} \quad
    Isaac Triguero\textsuperscript{1,2}\,\orcidlink{0000-0002-0150-0651} \quad
    Daniel Molina\textsuperscript{1,2}\,\orcidlink{0000-0002-4175-2204}\\
    \small\normalfont
    \textsuperscript{1}Department of Computer Science and Artificial Intelligence, School of Computer Science and Telecommunications Engineering (ETSIIT), University of Granada, 18071 Granada, Spain \\
    \textsuperscript{2}Andalusian Research Institute in Data Science and Computational Intelligence, 18016 Granada, Spain \\
    \textsuperscript{3}Lab for Uncertainty in Data and Decision Making (LUCID), School of Computer Science, University of Nottingham, NG7 2RD Nottingham, UK
}
\date{}

\renewcommand{\shorttitle}{Too much of a good thing: Block-wise KD and overfitting}

\hypersetup{
pdftitle={Too much of a good thing -- when knowledge distillation promotes overfitting, and how to avoid it},
pdfsubject={cs.LG, cs.CV},
pdfauthor={Irene Trigueros-Lorca, Leonardo Concepción, Christian Wagner, Isaac Triguero, Daniel Molina},
pdfkeywords={Deep Learning, Knowledge Distillation, Convolutional Neural Networks, Block-wise Distillation, Data Scarcity, Fine-grained Classification},
}

\begin{document}
\maketitle

\begin{abstract}
The growing size of Convolutional Neural Networks has led to increasingly large and costly models. 
Knowledge Distillation (KD) addresses this by transferring knowledge from a large network (teacher) to a small one (student), also reducing the training data required.
KD is traditionally applied only at the network's final output. However, its behaviour when applied at intermediate network layers has received little attention. This raises the question of whether intermediate block-wise KD, which provides supervision throughout the network, could offer an advantage under specific conditions, such as few instances per class, which is common in fine-grained datasets.
This work proposes a student design based on simple, homogeneous blocks mirroring those of the teacher, distilling knowledge between corresponding blocks. 
Across eleven datasets, we show that on classic datasets, distilling only the last block is sufficient — and often best—, whereas fine-grained, data-scarce settings benefit substantially from intermediate supervision, with even a single additional distillation point narrowing the gap considerably. 
We further study how this supervision should be guided, exploring configurations of varying granularity and informed by an explainability analysis based on attention maps, Centered Kernel Alignment, and Grad-CAM, alongside the impact of teacher and student fine-tuning strategies. 
This work shows that intermediate block-wise distillation, guided appropriately, is key to building compact data-efficient models without sacrificing accuracy.
\end{abstract}


\section{Introduction}\label{sec:Introduction}

Deep Learning (DL)~\cite{Goodfellow-et-al-2016} has revolutionised Artificial Intelligence, offering powerful models capable of performing tasks not considered possible just a few years ago. 
Among its most impactful domains is image processing, where Convolutional Neural Networks (CNNs) have achieved strong performance in applications such as medical imaging~\cite{heDeepLearningbasedImage2025} or satellite imagery analysis~\cite{krejcarAIRemoteSensing2026}, producing a large and diverse family of models tailored to specific tasks and constraints~\cite{razaLightweightDeepLearning2025}. 
One reason for this versatility is that CNNs can be easily adapted to new problems by retaining their backbone and changing only the final layers, an adaptation often accelerated by Transfer Learning~\cite{Weiss2016}. 
However, fine-tuning these models for specific domains requires great computational power to train and deploy them due to their size~\cite{liuSurveyModelCompression2025}, motivating the need for smaller models.

The compression and acceleration of model training and inference is mainly tackled by four methods~\cite{liuSurveyModelCompression2025}: 1) quantization, 2) pruning, 3) decomposition, 4) knowledge distillation. 
The first three approaches act directly on the original network, optimizing its numerical representation, structure, or parameters to reduce computational cost during inference. 
Knowledge Distillation (KD)~\cite{mansourianComprehensiveSurveyKnowledge2025, gouKnowledgeDistillationSurvey2021}, in contrast, adopts a different strategy: it transfers the learned knowledge into a separate, more compact network, making it particularly suitable when architectural flexibility or task specialization is required. 
One of the most popular KD methods is the teacher-student architecture~\cite{huTeacherStudentArchitectureKnowledge2023}. 
This methodology consists of two neural networks: a large pre-trained model (teacher) and a smaller model (student), which is trained under the guidance of the teacher's output. 
It has been shown that this knowledge transfer during training can greatly improve the process, achieving very competitive results with the smaller model. 
It is also recognized that KD allows more efficient learning on datasets with scarce data~\cite{cheraghianSemanticawareKnowledgeDistillation2021, liuLowresolutionFewshotLearning2024}. 
In particular, fine-grained datasets often face data scarcity due to the high cost of expert annotation required to label classes with subtle visual differences. 

While most KD methods transfer knowledge through the teacher's final output predictions, feature-based methods instead transfer the internal representations of the teacher network, aiming to provide richer and more granular supervision signals that capture how the teacher builds its representations progressively, rather than just its final decision. 
This could be particularly relevant in data-scarce scenarios, where additional supervision signals along the network might help compensate for the limited availability of training examples. 
Furthermore, the modular structure of many popular CNN models, such as ResNet, VGG, or EfficientNet, composed of repeated homogeneous blocks of increasing complexity, could lend itself to this type of intermediate distillation, offering an additional avenue for more effective and structured knowledge transfer across the network.


However, despite its potential, intermediate feature-based distillation remains comparatively less explored. 
Although indicated as a possibility since the beginning~\cite{hintonDistillingKnowledgeNeural2015}, most proposals apply it only at the end of the model~\cite{alkhulaifiKnowledgeDistillationDeep2021b} or the end of the convolutional part~\cite{gaoEmbarrassinglySimpleApproach2019a}. 
This lack of attention is compounded by several open challenges that intermediate KD still faces. 
There is still no consensus on which layers or blocks should be selected for distillation, nor on whether intermediate distillation points are worth introducing and, if so, how many should be applied~\cite{mansourianComprehensiveSurveyKnowledge2025}. 
In fact, too few points may add little over end-point distillation, while too many may introduce redundant or conflicting supervision signals. 
Beyond these design questions, data scarcity itself remains comparatively underexplored from this angle: existing efforts have mostly relied on few-shot or data-free formulations, rather than examining how the distillation depth interacts with dataset complexity and size. 
Additionally, the diversity of existing proposals regarding what type of representation to transfer, ranging from raw activations to attention maps~\cite{zagoruykoPayingMoreAttention2017}, contrastive~\cite{tianContrastiveRepresentationDistillation2020}, or relational knowledge~\cite{parkRelationalKnowledgeDistillation2019}, further reflects the lack of agreement on how intermediate knowledge should be extracted and distilled. 


In this work, in order to analyse the relationship between data-scarcity and the KD process, we propose a design for a student model composed of repetitive homogeneous blocks, using the same number of blocks as the original CNN teacher model, but with a simpler internal structure.
We introduce KD between corresponding blocks of the teacher and student models, where the student is trained to match the teacher's output feature maps at the end of each corresponding block. 
The first contribution of this paper is a systematic study of the effect of applying KD at different numbers of intermediate blocks, demonstrating that block-wise intermediate distillation substantially enhances learning efficiency, particularly in data-scarce and fine-grained settings. 
The second contribution is a diagnostic-then-guided design analysis of KD across intermediate blocks, using explainability to identify the conditions under which intermediate distillation provides the greatest benefit. 
The main contributions of this work can be divided into the following objectives:
\begin{enumerate}[label=\textbf{RQ\arabic*}]
    \item To analyse the impact of different fine-tuning strategies within a block-wise KD framework, comparing fine-tuning the teacher alone, the student alone, and both jointly — a relevant consideration in practical scenarios where fine-tuning the teacher is not always possible, whether due to limited access or limited computational resources. 
    \item To evaluate in which scenarios applying knowledge distillation at intermediate blocks achieves better results than end-point distillation, with particular focus on data-scarce and fine-grained classification settings. 
    \item To investigate the role of data scarcity and fine-grained complexity as independent factors in block-wise KD, through controlled experiments on data augmentation and data reduction strategies.
    \item To explore the effect of the number of intermediate distillation points through evenly distributed distillation points.
    \item To examine how knowledge is effectively transferred across intermediate blocks using explainability techniques, identifying which blocks contribute most to effective distillation. 
    \item To study intermediate block-wise distillation configurations motivated by the explainability analysis. 
\end{enumerate}






The remainder of this contribution is organized as follows. 
In \autoref{sec:RelatedWork}, we review existing work on KD and explainability in KD. 
In \autoref{sec:Methodology}, we present the proposed block-wise KD framework and the explainability techniques used to analyse it.  
In \autoref{sec:Framework}, we describe the experimental setup, including datasets, models, and training details. 
In \autoref{sec:Experimental}, we present and discuss the results obtained across the different experimental studies. 
Finally, in \autoref{sec:Conclusions}, we summarise the main findings and outline future work. 

\section{Related Work}\label{sec:RelatedWork}

\subsection{Knowledge Distillation}

Knowledge Distillation methods can be broadly categorised into logit-based, feature-based, and similarity-based. 
Logit-based methods train the student to match the teacher's output distribution. 
The seminal work of Hinton et al.~\cite{hintonDistillingKnowledgeNeural2015} introduced KD by matching softened class probabilities using KL divergence, exploiting the dark knowledge contained in non-target class scores. 
Subsequent work has refined the quality of this signal. 
DKD~\cite{zhaoDecoupledKnowledgeDistillation2022} decouples target-class and non-target-class knowledge, while DIST~\cite{huangKnowledgeDistillationStronger2022} preserves the relational structure among logits. 
More recently, some works questioned whether the teacher's raw output distribution is the optimal supervision signal, proposing to reshape  it via energy-based formulations~\cite{kimMaximizingDiscriminationCapability2024} or more compact representations~\cite{yuanStudentfriendlyKnowledgeDistillation2024} before distillation. 

While these methods improve how the teacher's decision is exploited, they remain limited to its final output, disregarding how that decision is built internally across the network. 
Feature-based methods transfer knowledge through intermediate representations within the network, introducing supervision at different depths. 
A key design choice is the location of distillation, i.e., which layers or structures are used to transfer the knowledge.
FitNet~\cite{romeroFitNetsHintsThin2015a} introduces hint-based loss at a single intermediate layer. 
Factor Transfer~\cite{kimParaphrasingComplexNetwork2018} applies distillation at deep representation, typically the final convolutional stage before pooling, introducing a paraphraser to extract compact teacher factors and a translator to align student features. 
Other approaches extend supervision across multiple layers, including AT~\cite{zagoruykoPayingMoreAttention2017}, which matches attention maps at several depths, and ReviewKD~\cite{chenDistillingKnowledgeKnowledge2021}, which aggregates and refines information from multiple intermediate representations to improve feature alignment. 
Block-structured distillation methods, such as DNA~\cite{liBlockWiselySupervisedNeural2020a}, operate on grouped representations rather than individual layers but require a complex and time-consuming optimization process to design a heterogeneous block structure. 
Similarly, CBKD~\cite{lanCounterclockwiseBlockbyblockKnowledge2025} performs distillation sequentially from deeper to shallower blocks, progressively replacing teacher blocks under channel-reduction constraints. 


These methods, however, transfer representations independently for each sample, without considering how samples relate to one another. 
Similarity-based distillation methods instead preserve the relational structure of the teacher's representation space. 
RKD~\cite{parkRelationalKnowledgeDistillation2019} distils pairwise distances and angular relationships between samples, replicating the geometric structure of the teacher embedding space. 
Contrastive approaches extend this by explicitly separating positive and negative relationships.  
CRD~\cite{tianContrastiveRepresentationDistillation2020} maximises agreement between teacher and student representations while contrasting against negative examples. 
More recently, BicKD~\cite{zhuBicKDBilateralContrastive2026} introduces a bilateral, orthogonality-enforcing contrastive loss over both sample-wise and class-wise relationships. 

Our work falls within the feature-based category, focusing specifically on block-wise distillation. 
Unlike single-layer or single-stage approaches, we distil knowledge across multiple blocks of a fixed, homogeneous student architecture mirroring the teacher's structure, enabling a systematic study of how the number and location of distillation points affect performance across different dataset complexities and data availability conditions. 

\subsection{Explainability in KD}

Understanding what knowledge is actually transferred during distillation, and how it propagates through the student network, remains an  open question that accuracy alone cannot answer. 
Explainability methods — particularly saliency maps and class activation maps (CAMs) — have increasingly been used to address this, both as distillation signals and as post-hoc analytical tools. 
The concept of using spatial activation maps to align teacher and student was established by AT~\cite{zagoruykoPayingMoreAttention2017}, already discussed in the context of feature-based distillation, which compresses intermediate feature maps into 2D attention heatmaps and minimises their L2 distance. 
Building on this direction, CAT~\cite{guo_class_2023} operates at the logit level, normalising CAMs by class and minimising their MSE, while 
Exp-KD~\cite{sun_explainability-based_2025} generates CAMS for multiple top-K predicted classes simultaneously, avoiding the need for backpropagation. 
Grad-CAM~\cite{selvaraju_grad-cam_2020} has also been 
used to guide distillation, most notably by e$^2$KD~\cite{parchami-araghi_good_2025}, which combines KL divergence on logits with cosine similarity on Grad-CAM or B-cos explanations in a model-agnostic loss, demonstrating particular benefits under distribution shift and in data-scarce regimes. 
Beyond activation-based approaches, Centered Kernel Alignment (CKA)~\cite{kornblith_similarity_2019}, a metric originally proposed for quantifying representation similarity, has also been used to guide training directly. 
Zhou et al.~\cite{zhouRethinkingCenteredKernel2024} show that CKA computes the cosine similarity between Gram matrices, and that maximising it is equivalent to minimising an upper bound of the Maximum Mean Discrepancy, justifying its use as a training objective. 

On the post-hoc analysis side, Cheng et al.~\cite{cheng_explaining_2020} propose three quantitative metrics, 
showing that distilled students learn more foreground-relevant features more efficiently than models trained from scratch, although these require bounding box annotations and do not correlate directly with accuracy. 
Similarly, UniCAM~\cite{adhane_explaining_2025} introduces a Grad-CAM variant based on partial distance correlation to isolate features uniquely transferred from the teacher. 
From a more analytical perspective, Stanton et al.~\cite{stanton_does_2021}
show that high accuracy does not imply that the student closely matches the teacher's behaviour,
and that neither more data nor better optimisers resolve the underlying difficulty. 
A mechanistic explanation is proposed by Ojha et al.~\cite{ojha_what_2023}, showing that regardless of the correctness of individual CAMs, distilled students converge to activation maps similar to their teacher's across logit-based, feature-based, and contrastive distillation alike, an effect attributed to the geometric consequence of mimicking the teacher's decision boundary. 
Together, these findings suggest that verifying the student's interpretability — beyond its accuracy — is an important step when proposing new distillation strategies. 

We adopt a similar post-hoc perspective, using attention maps, CKA, and Grad-CAM alike as diagnostic tools to examine how knowledge is distilled across intermediate blocks during distillation, rather than to guide the training process itself. 

\subsection{Data-Efficient and Fine-Grained Settings}

Data scarcity is a common challenge in many real-world applications, where collecting large labelled datasets is often impractical due to cost, time, or domain-specific constraints. 
KD has been shown to be particularly effective in such low-data regimes, for instance in few-shot class-incremental learning~\cite{cheraghianSemanticawareKnowledgeDistillation2021} or low-resolution few-shot classification~\cite{liuLowresolutionFewshotLearning2024}.
Lanzillotta et al.~\cite{lanzillotta_revisiting_2025} further demonstrate that the effect of distillation is not only preserved but amplified as dataset size decreases, coining this the data efficiency of distillation. 
Existing work addressing data scarcity in KD typically does so through few-shot or data-free~\cite{liu_mosaic_2026} formulations; 
meanwhile, performance on fine-grained datasets has received comparatively little attention as a desirable characteristic for feature distillation methods. 

Our work directly addresses this gap by evaluating block-wise distillation on fine-grained, data-scarce datasets, and by explicitly disentangling the two factors through controlled data augmentation and data reduction experiments. 


\section{Block-wise Knowledge Distillation Using Homogeneous Blocks}
\label{sec:Methodology}

In this section, we provide a detailed description of our proposal, beginning with the overall scheme, followed by the architecture of the student block, the training process, the flexible application of KD across block configurations, and the explainability techniques used to analyse model behaviour. 


\subsection{Global Scheme}

Following the usual architectures of popular CNNs, the teacher model is divided into blocks of consecutive layers with the same architecture. 
In order to apply KD to each block, the student is also composed of the same number of blocks. 
In our case, since the teacher is EfficientNet-B0, the model is naturally divided into six blocks, which can be merged into coarse groups to reduce the number of distillation points. 
The models analysed in this work follow a block-wise KD methodology in which the final layers (classification layers) are excluded from the KD process. 

\subsection{Student's block Structure}

Each block in the student is composed of two Inverted Residual modules~\cite{sandlerMobileNetV2InvertedResiduals2018}. 
Each module is composed of a channel expansion via $1\times1$ convolution with an expansion ratio of 6,  followed by normalisation, a $7\times7$ depthwise convolution with normalisation, and a Squeeze-and-Excitation~\cite{huSqueezeandExcitationNetworks2018} module that reduces the channel dimension by a factor of 24 before expanding it back. 
Finally, a $1\times1$ convolution projects the features back to the module's output channel dimension, followed by normalisation; when input and output shapes match, a residual skip connection is added. 
The first module of each block applies a stride of 2 for spatial downsampling, except in blocks 3 and 5, which retain a stride of 1, following the same downsampling schedule as the teacher architecture. 
The second module of each block always uses a stride of 1. 
The output channel dimension of each block is set to match that of its corresponding teacher block. 



\subsection{Training Process}
\label{sec:training}

\begin{figure*}[htp]
    \centering
    \includegraphics[width=\textwidth]{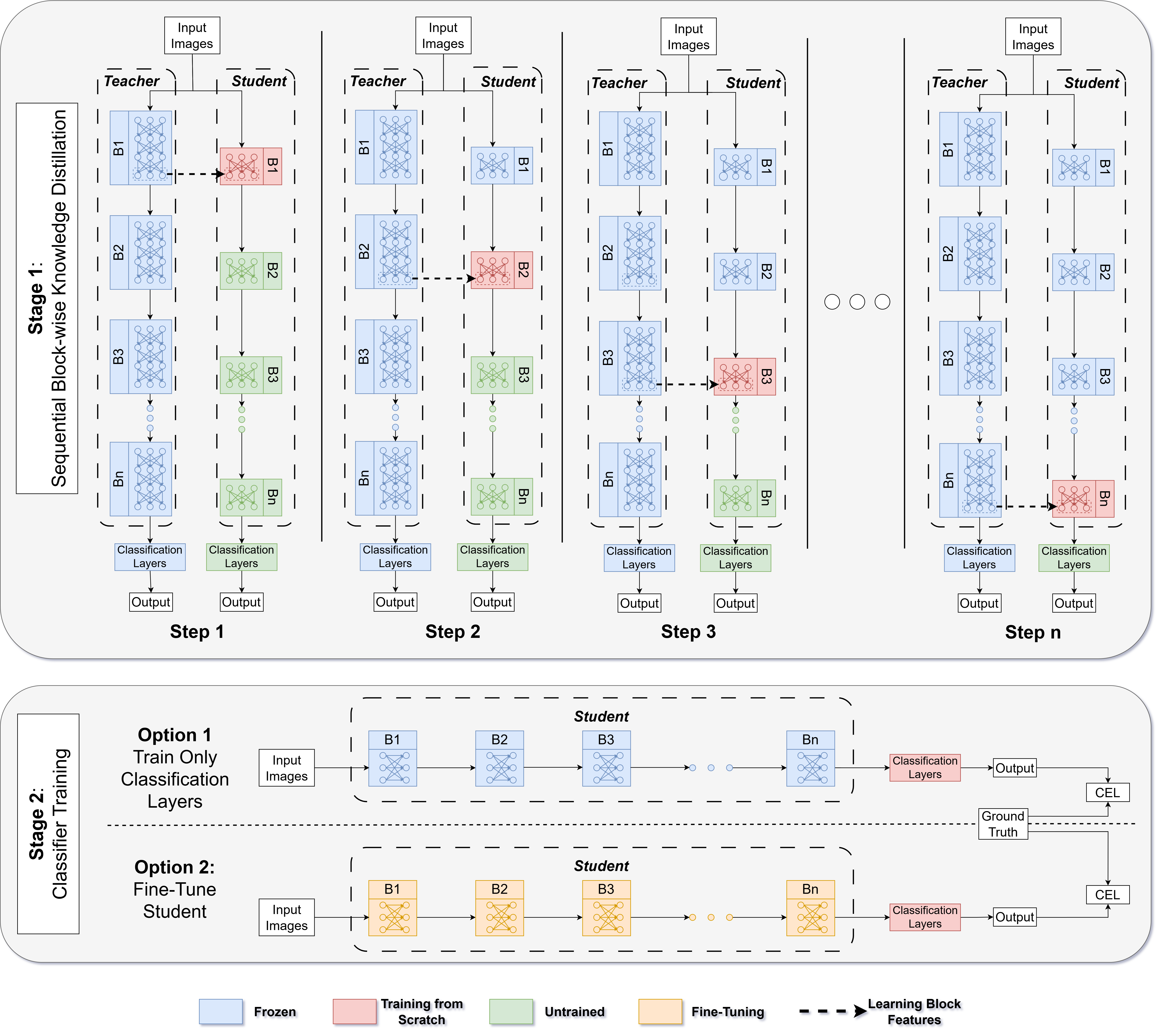}
    \caption{Illustration of our progressive block-wise Knowledge Distillation scheme}
    \label{fig:GeneralProcess}
\end{figure*}

In this strategy, we use a pre-trained model as the teacher, which may or may not be fine-tuned on the specific dataset. 
The student's training consists of two stages: training all its blocks via KD and training the classification layers. 
This process is illustrated in  Fig.~\ref{fig:GeneralProcess}.

In the first stage, the student's blocks are trained sequentially via KD, with each block's output computed only once its turn to be trained is reached. 
The objective of this stage is for each student block to replicate the corresponding teacher block's output, ignoring the classification entirely, which is addressed in the second stage. 
The loss function used for distillation is the Mean Squared Error (MSE) between the output feature maps of the corresponding student and teacher blocks. 
While a block is being trained, its input is the teacher's feature map at the preceding block, not the student's so no gradient reaches any other part of the student; other blocks keep their current weights. 
Each block undergoes training for a fixed number of 20 epochs. 

The number of distillation points can also be reduced by treating several consecutive blocks as a single unit, a configuration described in more detail in \autoref{sec:kd_levels}. 
In such case, only the last block of the group receives a direct KD supervision from the teacher, the remaining blocks within that same group are updated just by backpropagation. 

In the second stage, the student's classification layers, whose weights are still randomly initialised and have not been involved in the KD process, are trained to address the classification problem itself. 
Two configurations are considered for this stage: in the first, all of the student's blocks remain frozen and only the classification layers are trained; in the second, the entire student is left unfrozen, allowing the training of the classification layers to also fine-tune the rest of the student. 
In both cases, training is constrained to a maximum of 300 epochs, applying an early stopping criterion with a patience of 15 epochs, and enforcing a minimum of 100 epochs to avoid stopping at an early,  potentially premature stage. 
The final model retained is the one corresponding to the epoch with the best validation performance, rather than the one obtained at the last training epoch. 

\subsection{Different applications of KD to blocks}
\label{sec:kd_levels}

\begin{figure*}[ht]
    \centering
    \includegraphics[width=\linewidth]{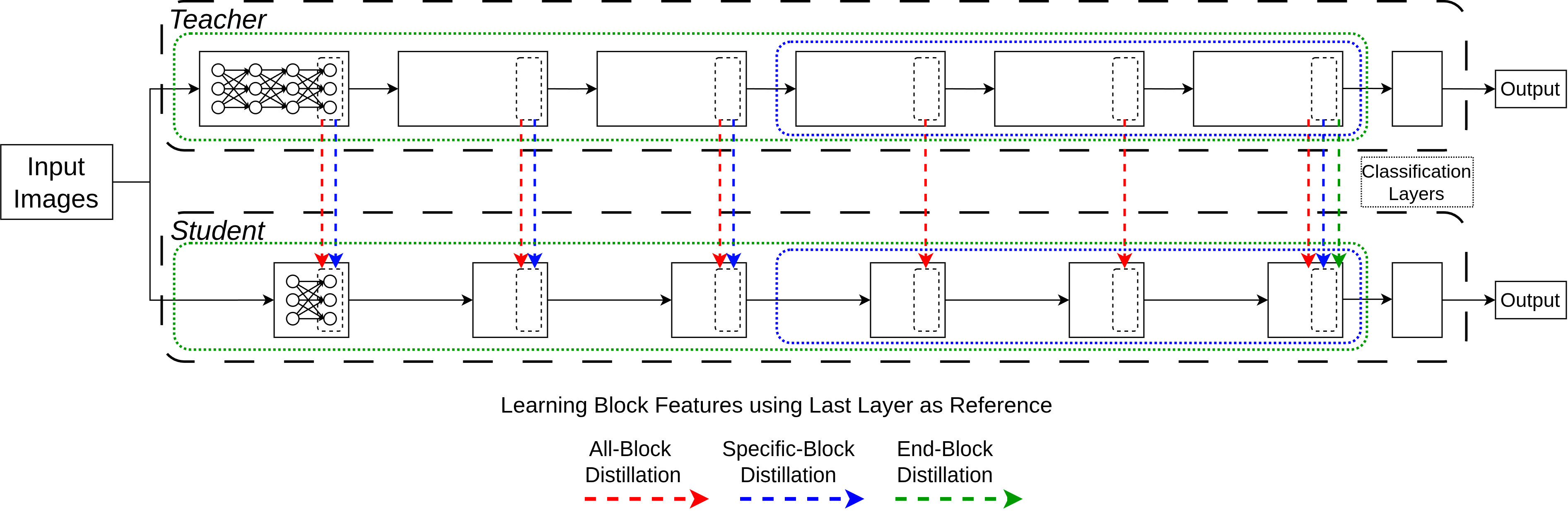}
    \caption{Illustration of the flexibility of the proposed framework, showcasing different Knowledge Distillation configurations. Marked in red, KD is applied to all blocks. In blue, KD is applied to a specific subset of blocks (1, 2, 3, and 6). In green, KD is applied only to the last block.}
    \label{fig:kd_levels}
\end{figure*}

The block-wise KD approach presented in this work offers a flexible framework that can accommodate different KD configurations, depending on how many and which blocks of the student are directly distilled, as illustrated in Fig.~\ref{fig:kd_levels}. 

Applying KD to all blocks, as shown in Fig.~\ref{fig:kd_levels}-All-Block, corresponds to the configuration with the highest number of distillation points, 
and is therefore expected to yield the greatest feature-wise similarity between teacher and student representations, since every block receives an explicit distillation signal. 
At the opposite end, KD can be restricted to the final block only, with the remaining blocks trained purely through backpropagation, as shown in Fig.~\ref{fig:kd_levels}-End-Block; this corresponds to a commonly adopted scheme in which only the teacher's final embedding is taken into account. 
Between these two extremes, intermediate configurations can be defined, such as the example in Fig.~\ref{fig:kd_levels}-Specific-Block, where KD is applied to a specific subset of blocks (e.g.,  blocks 1, 2, 3, and 6).

Since the number of layers and the overall complexity of the student network remain unchanged across configurations, any observed differences in performance appear to depend on which blocks were trained with KD. 


\subsection{XAI}
\label{sec:XAI_method}

To better understand how knowledge is distilled through the student network and to identify which blocks encode more discriminative information, we employ three complementary explainability techniques: attention maps, CKA, and Grad-CAM. 
Each method is applied at the output of every block of the network, regardless of whether that block is used as a distillation point, allowing us to examine whether and how the teacher's information is recovered even at blocks without direct distillation supervision, rather than being restricted to the final layer as in most conventional explainability approaches. 

As a first and most direct approach, we derive attention maps from the feature maps produced at each block's output. 
Given a feature map $F_j\in\mathbb{R}^{C\times H \times W}$ at block $j$, with $C$ channels and spatial dimensions $H\times W$, the corresponding attention map $A_j\in\mathbb{R}^{H\times W}$ is obtained by squaring the activations and summing across the channel dimensions, with $(x,y)$ indexing spatial position:
\begin{equation}
    A_j(x,y) = \sum_{c=1}^C F_j(c,x,y)^2
\end{equation}
This operation compresses the channel-wise information into a single spatial map that highlights the regions eliciting the strongest activations in that block, allowing a rationale similar to activation-based attention transfer methods~\cite{zagoruykoPayingMoreAttention2017}. 
To compare teacher and student attention maps for corresponding blocks, we use cosine similarity, which was found to yield consistent conclusions during preliminary tests. 

While attention maps offer a direct, pixel-level comparison, we complement this analysis with CKA~\cite{kornblith_similarity_2019}, a similarity metric that quantifies how comparable two sets of network representations are, even when they differ in dimensionality. 
Intuitively, CKA 
compares not individual activations directly, but whether the relative structure of these representations — i.e., how pairs of examples relate to one another within each representation — is preserved across both. 
In our setting, this allows us to estimate, for a given pair of teacher-student blocks, how much of the teacher's block information is present in the student's. 
Given the activations of two layers (or blocks) with $p_1$ and $p_2$ neurons (via global average pooling), respectively, collected over the same set of $n$ input samples, represented as matrices $X\in\mathbb{R}^{n\times p_1}$ and $Y\in\mathbb{R}^{n\times p_2}$, CKA computes the Hilbert-Schmidt Independence Criterion (HSIC) between their Gram matrices $K=XX^T$ and $L=YY^T$ via an unbiased minibatch estimator, normalised to produce a similarity score bounded between 0 and 1: 
\begin{equation}
    \text{CKA}(K,L) = \dfrac{\text{HSIC}(K,L)}{\sqrt{\text{HSIC}(K,K)\cdot\text{HSIC}(L,L)}}
\end{equation}
Zhou et al.~\cite{zhouRethinkingCenteredKernel2024} show that this formulation is equivalent to computing the cosine similarity between the Gram matrices $K$ and $L$, offering a more intuitive geometric interpretation of what CKA measures beyond its kernel-based definition. 
A key property of CKA is its invariance to orthogonal transformations and isotropic scaling of the representations, making it suitable for comparing teacher and student features even when the underlying representational spaces are not directly aligned. 
This offers a more holistic perspective than the pixel-level attention maps described above. 

Finally, we employ Grad-CAM~\cite{selvaraju_grad-cam_2020}, which produces coarse localization maps by using the gradients of a target class score with respect to the feature maps of a chosen convolutional layer. 
Each channel of the feature map is weighted by the global average of its corresponding gradients, and a ReLU is applied to retain only the features that positively contribute to the target-class prediction. 
Unlike its predecessor, CAM~\cite{zhou_learning_2016}, which requires 
global average pooling immediately before the softmax layer, Grad-CAM does not impose any architectural constraint, since it relies solely on gradient information. 
This is relevant to our setting, allowing consistent application across all student blocks, not just its output. 
However, it is worth noting that, as shown by Selvaraju et al.~\cite{selvaraju_grad-cam_2020}, localisation quality tends to degrade at layers farther from the network's output, since earlier layers capture less semantic, more localized information due to their smaller receptive fields — a factor that is particularly relevant in our setting, as it may affect the reliability of Grad-CAM when applied to early blocks.

\section{Experimental Framework}
\label{sec:Framework}

For our experiments, we have selected EfficientNet-B0~\cite{tanEfficientNetRethinkingModel2019} as the teacher, a well-known CNN model for its performance and available directly in the DL libraries. 
The teacher has been pre-trained on ImageNet~\cite{cnnimagenet}, 
a standard choice for pre-training, 
and fine-tuned per dataset using SGD (learning rate 0.001, momentum 0.9) for up to 300 epochs, with early stopping (patience 15, minimum 30 epochs), retaining the checkpoint with the best validation accuracy. 

We have carried out the experiments with 11 classic image classification datasets, divided into seven classic datasets and four data-scarce, fine-grained ones.
Their properties are summarised in \autoref{tbl:datasets}, covering number of examples, classes and examples per class. 
For several imbalanced datasets, that ratio is an interval, since it depends on the class. 
SVHN is so imbalanced that any value could be confusing.

\addtolength{\tabcolsep}{-.38em}

\begin{table*}[htp]
\addtolength{\tabcolsep}{-.12em}
\small
  \caption[Datasets]{Datasets used with their main features}
  \label{tbl:datasets}
  \centering
  \addtolength{\tabcolsep}{0.16em}
  \begin{tabular}{llrrrr|l}
    \toprule
    & Dataset & \(\#\)Training & \(\#\)Test & \(\#\)Classes & \(\frac{\#Training}{\#Classes}\) & Descriptions\\
    \midrule
 \multirow{7}{*}{\rotatebox[origin=c]{90}{\footnotesize \textbf{Classic Datasets}}}  &    CIFAR10~\cite{krizhevskyLearningMultipleLayers2009}  & 50,000  & 10,000 & 10  & 5,000 & 10-object categories image classification benchmark\\
     &    CIFAR100~\cite{krizhevskyLearningMultipleLayers2009} & 50,000  & 10,000 & 100 & 500 & 100-object categories image classification benchmark \\
     &    EMNIST~\cite{cohenEMNISTExtendingMNIST2017}          & 112,800 & 18,800 & 47  & 2,400 & Extended version of MNIST including handwritten letters \\
     &    FashionMNIST~\cite{xiaoFashionMNISTNovelImage2017}   & 60,000  & 10,000 & 10  & 6,000 & 10-class apparel image classification \\
     &    Food101\cite{bossardFood101MiningDiscriminative2014} & 75,750  & 25,250 & 101 & 750 & 101-class food image classification  \\
     &    MNIST~\cite{lecunGradientbasedLearningApplied1998}   & 60,000  & 10,000 & 10  & 6,000 & Handwritten digit recognition dataset \\
    &    SVHN~\cite{SVHN}                                     & 73,257  & 26,032 & 10  & - &  Digits in natural street-view images classification \\
    \midrule
    & & & & & \\
\multirow{4}{*}[0.7em]{\rotatebox[origin=c]{90}{\footnotesize \textbf{Data Scarce}}}
     &
     CUB200 ~\cite{wah2011cub2002011}                     & 5,994   & 5,794  & 200 & 30-60 & Bird species image classification\\
     &    ISIC~\cite{codellaSkinLesionAnalysis2018}            & 1,849   & 459    & 8   & ~220 & Skin lession image classification \\
     &    OxfordPets~\cite{parkhiCatsDogs2012c}                & 3,680   & 3,669  & 37  & 99 & Dog and cat breed image classification   \\
     &    StanfordCars~\cite{krause20133dobjects}            & 8,144   & 8,041  & 196 & 24-68 & Car model image classification\\
    \bottomrule
  \end{tabular}
\end{table*}

For each dataset, the training set has been further divided into an actual training set (80\%) and a validation set (20\%) for early stopping. 
Following common practice, batch sizes have been adjusted according to dataset's characteristics: for ISIC it is 16; for the rest of fine-grained, Food101, and MNIST, 32; for the rest of datasets, 128. 
For the last stage of training, we have set the maximum number of epochs to 300, with an early stopping patience of 15. 
To avoid stopping at relative minima, we set a minimum of 100 epochs, 
training each model five times and reporting average test accuracy 
(median standard deviation across seeds was $0.2\%$, rising to \~{}$4\%$ only for a few fine-grained configurations). 

Inspired by \cite{liBlockWiselySupervisedNeural2020a}, but maintaining the homogeneity of the blocks, we have carried out a small manual experiment with a different number of Inverted Residual modules per block, from 1 to 4. 
Comparing all these models, we observed that using 2 yields results similar to those of the searched student using the complex search in that paper, while reducing the time needed to define and train it -- selected via manual comparison.
The rest of parameters used by the proposal are the ones established in \cite{liBlockWiselySupervisedNeural2020a}.

In this work, we study several KD configurations based on the subset of blocks selected for distillation, ranging from the two extreme configurations \fullnameUnoB and \fullnameSeisB, mentioned in \autoref{sec:kd_levels}, to a set of intermediate configurations explored through a granularity study and an explainability-guided study, presented in \autoref{sec:GranularityStudy} and \autoref{sec:ExplainabilityResults}, respectively. 
In the results tables, we highlight in bold the best accuracy value for each dataset, and underline the second best.


In some experiments, explicitly stated, we have applied Data Augmentation (DA). 
Using DA allowed each original image from the training subset to generate four variations, effectively increasing the subset's size by five while maintaining semantic consistency. 
The transformations selected 
were based on a literature review and with the aim of preserving class-distinctive features while introducing meaningful variability. 
For OxfordPets, we applied random horizontal flips, aggressive colour jittering, random grayscale, and Gaussian blur;  
for ISIC, domain-specific transformations including reflective padding, random affine transformations (rotation, shear, scale), bidirectional flips, and moderate colour jittering; 
and for CUB200 and StanfordCars,  more conservative augmentations (small rotations, horizontal flips, slight colour jittering) to avoid distorting discriminative features such as bird plumage patterns or vehicle design details. 

Similarly, in some experiments, we have applied Data Reduction via stratified random sampling: for each class, a fixed percentage of the total available samples was randomly selected, with the same percentage applied uniformly across all classes, preserving the original class distribution. 
The training and validation subsets described above were then obtained from this reduced set, following the same 80/20 split. 
We considered three reduction levels, retaining 5\%, 10\%, and 80\% of the original training data. 
Importantly, the distilled teacher model was also trained on the same reduced subset, to prevent data leakage from samples excluded by the reduction. 

We have implemented our architecture via PyTorch (2.6.0). 
The machine specifications used by our experiments include a node with two AMD EPYC 7742 64-Core Processors, 1 TiB  of RAM and 8 Nvidia A100 GPUs, each with 40 GB of VRAM.
However, each model has been trained on a single GPU at a time.

\section{Analysis of results}
\label{sec:Experimental}

In this section, we analyse the results obtained from different experimental studies. 
Specifically, our aims are:
\begin{itemize}
    \item To compare the different fine-tuning strategies against each other, assessing their relative impact on the resulting performance (\autoref{sec:FineTuning}).
    \item To assess how data availability affects intermediate KD, by independently augmenting fine-grained datasets and reducing classic ones (\autoref{sec:DataAvailability}).
    \item To examine whether evenly distributed intermediate configurations offer a favourable trade-off between the two extreme schemes (\autoref{sec:GranularityStudy}).
    \item To analyse, through explainability techniques, how information flows across blocks during distillation, and to derive new configurations from these insights (\autoref{sec:ExplainabilityResults}).
    \item To evaluate an explainability-guided student against the two extreme schemes (\autoref{sec:Blocks3456}).
\end{itemize}

\subsection{Influence of Fine-tuning in teacher and student models}
\label{sec:FineTuning}

In this section, we tackle the decision about whether to fine-tune the teacher or the student model. 
The compared algorithms and their notation are the following:

\begin{itemize}
    \item \textbf{Baseline:} These are the results of the teacher model (EfficientNetB0), pre-trained with ImageNet and fine-tuned for each problem.
    \item \textbf{Fine-tuned Teacher, FT:} The KD results using a fine-tuned teacher model, without fine-tuning the student model.
    \item \textbf{Fine-tuned Student, FS:} The KD results using a teacher model without fine-tuning, and later fine-tuning the student model.
    \item \textbf{Fine-tuned Teacher and Student, FTS:} The KD results using a fine-tuned teacher model, and later fine-tuning the student. 
\end{itemize}

For this analysis, we restrict our comparison to the two extreme block-wise configurations already introduced in \autoref{sec:kd_levels}: \fullnameUnoB (\nameUnoB), which distils knowledge only at the last block, and \fullnameSeisB (\nameSeisB), which distils knowledge at every block. 
These configurations represent the two boundary cases of block-wise distillation and therefore provide the clearest setting to isolate the effect of fine-tuning strategy.

\subsubsection{Fine-tuning Strategy Comparison}
\label{sec:FineTuning_byConfig}

The results obtained when applying KD only to the final block (\nameUnoB) are shown in \autoref{tbl:FineTuning_1B}, while those obtained when applying KD to all blocks (\nameSeisB) are presented in \autoref{tbl:FineTuning_6B}. The average processing time and complexity for both configurations are reported in \autoref{tbl:finetuned_time_1B6B}. 

\begin{table*}[ht]
\addtolength{\tabcolsep}{-.075em}
    \caption{Average accuracy according to the fine-tuning strategy for extreme KD configurations}
    \label{tbl:FT_1B_and_6B}
    \begin{subtable}{0.48\textwidth}
    \caption{\fullnameUnoB (\nameUnoB)}
    \label{tbl:FineTuning_1B}
    \addtolength{\tabcolsep}{3pt}
    \begin{tabular}{llrrrr}
         \toprule
         & Datasets     & \multicolumn{1}{c}{Baseline} & \multicolumn{1}{c}{FT} & \multicolumn{1}{c}{FS} & \multicolumn{1}{c}{FTS}   \\
         \midrule
         \multirow{7}{*}{\rotatebox[origin=c]{90}{\footnotesize \textbf{Classic Datasets}}}
         & CIFAR10      & \textbf{0.914}    & \textbf{0.914} & 0.873 & \underline{0.901} \\
         & CIFAR100     & \textbf{0.797}    & \underline{0.774} & 0.733 & 0.766 \\
         & EMNIST       & \underline{0.903}    & \textbf{0.907} & 0.897 & 0.898 \\
         & FashionMNIST & \underline{0.942}    & \textbf{0.944} & 0.936 & 0.937 \\
         & Food101      & \underline{0.803}    & \textbf{0.812} & 0.751 & 0.782 \\
         & MNIST        & \underline{0.995}    & \textbf{0.996} & \underline{0.995} & \textbf{0.996} \\
         & SVHN         & \underline{0.967}    & \textbf{0.972} & 0.957 & 0.965 \\
         \midrule
         &              &          &       &       &       \\
         \multirow{4}{*}[0.7em]{\rotatebox[origin=c]{90}{\footnotesize \textbf{Data Scarce}}}
         & CUB200       & \textbf{0.722}    & \underline{0.413} & 0.364 & 0.386 \\
         & ISIC         & \textbf{0.684}    & \underline{0.585} & 0.488 & 0.550 \\
         & OxfordPets   & \textbf{0.891}    & \underline{0.540} & 0.462 & 0.491 \\
         & StanfordCars & \textbf{0.696}    & 0.424 & \underline{0.449} & 0.409 \\
         \bottomrule
    \end{tabular}
    \end{subtable}
    \hfill
    \begin{subtable}{0.48\textwidth}
    \caption{\fullnameSeisB (\nameSeisB)}
    \label{tbl:FineTuning_6B}
    \addtolength{\tabcolsep}{3pt}
    \begin{tabular}{llrrrr}
        \toprule
         & Datasets     & \multicolumn{1}{c}{Baseline} & \multicolumn{1}{c}{FT} & \multicolumn{1}{c}{FS} & \multicolumn{1}{c}{FTS} \\
         \midrule
         \multirow{7}{*}{\rotatebox[origin=c]{90}{\footnotesize \textbf{Classic Datasets}}}
         & CIFAR10      & \textbf{0.914}    & \underline{0.898} & 0.889 & 0.897 \\
         & CIFAR100     & \textbf{0.797}    & 0.774 & 0.743 & \underline{0.778} \\
         & EMNIST       & \underline{0.903}    & \textbf{0.905} & 0.896 & 0.900 \\
         & FashionMNIST & \textbf{0.942}    & \underline{0.937} & 0.934 & 0.935 \\
         & Food101      & \textbf{0.803}    & \underline{0.784} & 0.734 & 0.775 \\
         & MNIST        & \textbf{0.995}    & \textbf{0.995} & \textbf{0.995} & \textbf{0.995} \\
         & SVHN         & \textbf{0.967}    & \underline{0.965} & 0.955 & \underline{0.965} \\
         \midrule
         &              &          &       &       &       \\
         \multirow{4}{*}[0.7em]{\rotatebox[origin=c]{90}{\footnotesize \textbf{Data Scarce}}}
         & CUB200       & \textbf{0.722}    & 0.672 & 0.659 & \underline{0.683} \\
         & ISIC         & 0.684    & \textbf{0.694} & 0.687 & \underline{0.691} \\
         & OxfordPets   & \textbf{0.891}    & \underline{0.845} & 0.797 & 0.819 \\
         & StanfordCars & 0.696    & 0.703 & \underline{0.744} & \textbf{0.746} \\
         \bottomrule
    \end{tabular}
    \end{subtable}
\end{table*}

\begin{table}[ht]
  \caption{Average processing time and complexity according to the fine-tuning strategy for extreme KD configurations}
  \label{tbl:finetuned_time_1B6B}
  \centering
  \addtolength{\tabcolsep}{0.8pt}
  \begin{tabular}{lccccc}
    \toprule
    Measure     & \textit{Baseline} & FT     & FS    & FTS    \\
    \midrule                       
    \#Parameters (M) & 4.020    & \textbf{2.707}  & \textbf{2.707} & \textbf{2.707}  \\
    Train. Time (\nameUnoB)  & 04h08m   & 05h15m & \textbf{02h02m}& 05h56m \\
    Train. Time (\nameSeisB) & 04h08m   & 06h27m & \textbf{02h57m}& 07h17m \\
    \bottomrule
  \end{tabular}
\end{table}

Across both configurations, fine-tuning only the teacher (FT) matches or outperforms fine-tuning both models (FTS), indicating that fine-tuning the student in addition to the teacher brings limited benefit that does not justify the computational cost. 
FT clearly outperforms fine-tuning only the student (FS), although, in this case, at a substantially higher computational cost, making FS preferable when resources are limited.

Compared to the baseline, \nameUnoB (FT) (\autoref{tbl:FineTuning_1B}) outperforms it in 6 of the 7 classic datasets, proving highly competitive with fewer parameters at the cost of 1 additional hour of training, but shows clear performance degradation on fine-grained datasets. 
\nameSeisB (FT) (\autoref{tbl:FineTuning_6B}), in contrast, remains competitive with the baseline across all datasets despite using fewer parameters — the baseline is only marginally superior on 7 of the 11 — and shows no degradation on fine-grained datasets. 

Based on this analysis, for the remainder of this work, we focus on two fine-tuning options: FT, which achieves the best results, and FS, which achieves competitive performance at a lower computational cost. 

\subsubsection{End-Block vs All-Block Distillation}
\label{sec:FineTuning_Comparison}

In this section, we compare \fullnameUnoB and \fullnameSeisB models in more detail. 
The results obtained are shown in \autoref{tbl:FineTune_1B_vs_6B}, and their average processing times and complexities are reported in \autoref{tbl:finetuned_time_1B6B}.
Although these results can also be derived from \autoref{tbl:FT_1B_and_6B}, we present them here to facilitate direct comparison between the two configurations. 

\begin{table}[ht]
\centering
\caption{Average accuracy comparison of the baseline, KD applied only to the last block (\nameUnoB), and KD applied to all blocks (\nameSeisB)}
\label{tbl:FineTune_1B_vs_6B}
\addtolength{\tabcolsep}{1.2pt}
\begin{tabular}{llrrrrr}
\cmidrule{4-7}
 &              & \multicolumn{1}{l}{} & \multicolumn{2}{c}{\nameUnoB} & \multicolumn{2}{c}{\nameSeisB}   \\
 \toprule
 & Datasets     & \multicolumn{1}{c}{Baseline} & \multicolumn{1}{c}{FT} & \multicolumn{1}{c}{FS} & \multicolumn{1}{c}{FT} & \multicolumn{1}{c}{FS} \\
 \midrule
 \multirow{7}{*}{\rotatebox[origin=c]{90}{\footnotesize \textbf{Classic Datasets}}}
 & CIFAR10      & \textbf{0.914}       & \textbf{0.914}     & 0.873    & \underline{0.898}          & 0.889          \\
 & CIFAR100     & \textbf{0.797}       & \underline{0.774}              & 0.733    & \underline{0.774}          & 0.743          \\
 & EMNIST       & 0.903                & \textbf{0.907}     & 0.897    & \underline{0.905}          & 0.896          \\
 & FashionMNIST & \underline{0.942}                & \textbf{0.944}     & 0.936    & 0.937          & 0.934          \\
 & Food101      & \underline{0.803}                & \textbf{0.812}     & 0.751    & 0.784          & 0.734          \\
 & MNIST        & \underline{0.995}                & \textbf{0.996}     & \underline{0.995}    & \underline{0.995}          & \underline{0.995}          \\
 & SVHN         & \underline{0.967}                & \textbf{0.972}     & 0.957    & 0.965          & 0.955          \\
 \midrule
 &              &                      &                    &          &                &                \\
 \multirow{4}{*}[0.7em]{\rotatebox[origin=c]{90}{\footnotesize \textbf{Data Scarce}}}
 & CUB200       & \textbf{0.722}       & 0.413              & 0.364    & \underline{0.672}          & 0.659          \\
 & ISIC         & 0.684                & 0.585              & 0.488    & \textbf{0.694} & \underline{0.687}          \\
 & OxfordPets   & \textbf{0.891}       & 0.540              & 0.462    & \underline{0.845}          & 0.797          \\
 & StanfordCars & 0.696                & 0.424              & 0.449    & \underline{0.703}          & \textbf{0.744} \\
 \bottomrule
\end{tabular}
\end{table}

On classic datasets, \nameUnoB achieves the best performance, though \nameSeisB remains competitive, typically only 1-3\% behind in absolute accuracy. 

In contrast, on fine-grained datasets, \nameSeisB substantially outperforms \nameUnoB, with differences considerably larger than in the classic setting. 
This suggests that relying exclusively on the deepest block is associated with a severe loss of the fine-grained discriminative information that intermediate blocks preserve. 
Notably, \nameSeisB also narrows the gap with the baseline considerably in this regime, occasionally even surpassing it, while \nameUnoB remains far below the baseline across all four datasets. 
Regarding processing time (\autoref{tbl:finetuned_time_1B6B}), training \nameUnoB requires, on average, approximately one hour less than training \nameSeisB.

\subsection{Effects of Data Quantity}
\label{sec:DataAvailability}

The results discussed above suggest that applying KD to all blocks outperforms applying it only to the last block on datasets with few instances per class. 
In this section, we investigate whether the number of training instances is indeed the underlying factor. 
To this end, we manipulate the data quantity in both directions.
First, we apply Data Augmentation (DA) to fine-grained datasets to assess whether increasing the number of instances narrows the gap between \nameUnoB and \nameSeisB. 
We also reduce the amount of training data on classic datasets, to assess whether this induces the opposite effect. 
DA is applied exclusively to fine-grained datasets due to the substantial increase in training time. 

\subsubsection{Data Augmentation on Fine-Grained Datasets}
\label{sec:DataAugmentation}

In this section, we apply Data Augmentation (DA) to fine-grained datasets and compare the resulting accuracy against the non-augmented models discussed in \autoref{sec:FineTuning}. 
\autoref{tbl:DA_1B} and \autoref{tbl:DA_6B} show the results of applying DA to \nameUnoB and \nameSeisB models, respectively, alongside the corresponding average processing times.

\begin{table*}[ht]
\addtolength{\tabcolsep}{-.15em}
    \caption{Average accuracy and processing time with and without Data Augmentation (DA) on fine-grained datasets}
    \label{tbl:DataAugmentation}
    \begin{subtable}{0.48\textwidth}
    \caption{\fullnameUnoB (\nameUnoB)}
    \label{tbl:DA_1B}
    \addtolength{\tabcolsep}{3.5pt}
    \begin{tabular}{lrrrr}
        \toprule
        Datasets     & \multicolumn{1}{c}{FT} & \multicolumn{1}{c}{FT\_DA} & \multicolumn{1}{c}{FS} & \multicolumn{1}{c}{FS\_DA} \\
        \midrule
        CUB200       & 0.413                  & {\ul 0.646}                & 0.364                  & \textbf{0.664}             \\
        ISIC         & 0.585                  & \textbf{0.721}             & 0.488                  & {\ul 0.704}                \\
        OxfordPets   & 0.540                  & \textbf{0.799}             & 0.462                  & {\ul 0.757}                \\
        StanfordCars & 0.424                  & {\ul 0.729}                & 0.449                  & \textbf{0.732}             \\
        \midrule
        \midrule
        Train. Time &  00h42m & 05h03m & 01h12m & 04h42m \\
        \bottomrule
    \end{tabular}
    \end{subtable}
    \hfill
    \begin{subtable}{0.48\textwidth}
    \caption{\fullnameSeisB (\nameSeisB)}
    \label{tbl:DA_6B}
    \addtolength{\tabcolsep}{3.5pt}
    \begin{tabular}{lrrrr}
        \toprule
        Datasets     & \multicolumn{1}{c}{FT} & \multicolumn{1}{c}{FT\_DA} & \multicolumn{1}{c}{FS} & \multicolumn{1}{c}{FS\_DA} \\
        \midrule
        CUB200       & {\ul 0.672}            & \textbf{0.697}             & 0.659                  & 0.643                      \\
        ISIC         & 0.694                  & {\ul 0.717}                & 0.687                  & \textbf{0.736}             \\
        OxfordPets   & {\ul 0.845}            & \textbf{0.873}             & 0.797                  & 0.841                      \\
        StanfordCars & 0.703                  & 0.725                      & {\ul 0.744}            & \textbf{0.752}            \\
        \midrule
        \midrule
        Train. Time  & 01h23m & 07h35m & 00h47m & 07h08m \\
        \bottomrule
    \end{tabular}
    \end{subtable}
\end{table*}


In \autoref{tbl:DataAugmentation}, we observe that applying DA allows \nameUnoB models to narrow the performance gap with \nameSeisB models. 
This suggests that the superior performance of \nameSeisB over \nameUnoB on fine-grained problems may be related to the limited number of instances per class in these datasets, and that intermediate knowledge transferred by \nameSeisB may be particularly relevant in such scenarios. 

Although incorporating DA into \nameUnoB reduces these differences, applying it to \nameSeisB models also improves their results (as shown in \autoref{tbl:DA_6B}), indicating that these KD models also benefit from DA, although to a smaller extent than \nameUnoB models. 
This is consistent with \nameSeisB already leveraging intermediate blocks to compensate for limited data availability. 
However, DA increases processing time by a factor of four in the best case, which may not be a viable option in many scenarios. 

\subsubsection{Data Reduction on Classic Datasets}
\label{sec:DataReduction}

In this section, we reduce the amount of training data available on classic datasets. 
\autoref{tbl:DataReduction} shows the accuracy across four data availability percentages, including the full-data setting already discussed in \autoref{sec:FineTuning}.

\begin{table}[ht]
\addtolength{\tabcolsep}{0.3em}
\centering
\caption{Average accuracy with and without Data Reduction on classic datasets.}
\label{tbl:DataReduction}
\begin{tabular}{lrrrrrr}
\cmidrule{4-7}
                          &     &                       & \multicolumn{2}{c}{\nameUnoB}    & \multicolumn{2}{c}{\nameSeisB}   \\ 
\toprule
Datasets                  & \%Data  & \multicolumn{1}{c}{Baseline} & \multicolumn{1}{c}{FT} & \multicolumn{1}{c}{FS} & \multicolumn{1}{c}{FT} & \multicolumn{1}{c}{FS} \\ 
\midrule
\multirow{4}{*}{CIFAR10}  & 5   & 0.611                 & 0.535                  & 0.481                  & {\ul 0.624}            & \textbf{0.713}         \\
                          & 10  & 0.694                 & 0.650                  & 0.601                  & {\ul 0.710}            & \textbf{0.775}         \\
                          & 80  & 0.875                 & \textbf{0.889}         & 0.860                  & 0.869                  & {\ul 0.881}            \\
                          & 100 & {\ul 0.914}           & \textbf{0.914}         & 0.873                  & 0.898                  & 0.889                  \\ \hline
\multirow{4}{*}{CIFAR100} & 5   & \textbf{0.479}        & 0.184                  & 0.144                  & 0.344                  & {\ul 0.386}            \\
                          & 10  & \textbf{0.604}        & 0.290                  & 0.257                  & 0.486                  & {\ul 0.507}            \\
                          & 80  & \textbf{0.793}        & 0.734                  & 0.705                  & {\ul 0.750}            & 0.730                  \\
                          & 100 & \textbf{0.797}        & {\ul 0.774}            & 0.733                  & {\ul 0.774}            & 0.743                  \\ \hline
\multirow{4}{*}{Food101}  & 5   & \textbf{0.494}        & 0.164                  & 0.142                  & {\ul 0.490}            & 0.448                  \\
                          & 10  & \textbf{0.593}        & 0.347                  & 0.331                  & {\ul 0.587}            & 0.550                  \\
                          & 80  & \textbf{0.783}        & {\ul 0.782}            & 0.733                  & 0.760                  & 0.721                  \\
                          & 100 & {\ul 0.803}           & \textbf{0.812}         & 0.751                  & 0.784                  & 0.734                  \\ \hline
\multirow{4}{*}{SVHN}     & 5   & 0.683                 & 0.772                  & {\ul 0.831}                  & 0.746                  & \textbf{0.906}         \\
                          & 10  & 0.808                 & 0.861                  & {\ul 0.892}            & 0.831                  & \textbf{0.930}         \\
                          & 80  & 0.937                 & 0.951                  & {\ul 0.955}            & 0.938                  & \textbf{0.955}         \\
                          & 100 & {\ul 0.967}           & \textbf{0.970}         & 0.957                  & 0.965                  & 0.955                  \\
\bottomrule
\end{tabular}
\end{table}

As the percentage of available data decreases, \nameSeisB models, particularly under the FS strategy, increasingly outperform \nameUnoB models. 
This mirrors the pattern observed in \autoref{sec:DataAugmentation}, and further supports the hypothesis that the intermediate knowledge transferred by \nameSeisB becomes more valuable as the number of available instances per class decreases. 
Conversely, as the percentage of available data increases, \nameUnoB models, particularly under the FT strategy, become increasingly competitive, consistent with the results reported in \autoref{sec:FineTuning} for classic datasets with full data availability. 

These results point to a second factor beyond data availability: how closely the target domain matches the teacher's pre-training domain. 
When this domain gap is small, the teacher can rely on already-relevant pretrained features and remains difficult to surpass even under severe data reduction (as seen on CIFAR100 and Food101). 
When the gap is large, however, the teacher struggles to adapt the new features from limited fine-tuning data, allowing a model trained directly on the target domain to overtake it instead (as seen on SVHN).

Taken together, these results show that, for a small number of instances (10\% or lower), the performance of EBD models degrades significantly compared to the teacher. In contrast, ABD not only maintains highly competitive performance but also improves the teacher's results in several datasets, achieving surprisingly great results with only a small fraction of the data. This finding is particularly relevant for real-world problems, where data acquisition is often costly and limited data is the norm. Therefore, a model that achieves competitive results with few data points is highly valuable from a practical perspective.

\subsection{Granularity Study}
\label{sec:GranularityStudy}

The results discussed in the previous section reveal that \nameUnoB and \nameSeisB excel in different regimes: \nameUnoB on classic datasets, and \nameSeisB on fine-grained, data-scarce ones.  
Having evaluated these two extreme cases, this section investigates the effect of the number of blocks at which KD is applied. 
To this end, we conduct a granularity study, comparing the two extreme configurations discussed throughout this work against two naive intermediate configurations, obtained by evenly spacing the distillation points across the student's blocks: one using two blocks (Blocks36) and one using three blocks (Blocks246). 
\autoref{tbl:Granularity} shows the resulting accuracy for all four configurations, alongside the baseline. 

\begin{table}[ht]
\addtolength{\tabcolsep}{.3em}
\centering
\caption{Average accuracy for evenly spaced intermediate configurations, compared against \nameUnoB, \nameSeisB, and the baseline}
\label{tbl:Granularity}
\begin{tabular}{lccccc}
\toprule
Datasets     & Baseline       & \nameUnoB      & Blocks36            & Blocks246          & \nameSeisB  \\
\midrule
CIFAR10      & \textbf{0.914} & \textbf{0.914} & {\ul 0.913}    & {\ul 0.913}    & 0.898       \\
CIFAR100     & \textbf{0.797} & 0.774          & {\ul 0.792}    & \textbf{0.797} & 0.774       \\
EMNIST       & 0.903          & \textbf{0.907} & \textbf{0.907} & {\ul 0.906}    & 0.905       \\
FashionMNIST & {\ul 0.942}    & \textbf{0.944} & 0.941          & 0.940          & 0.937       \\
Food101      & 0.803          & \textbf{0.812} & {\ul 0.809}    & 0.801          & 0.784       \\
MNIST        & {\ul 0.995}    & \textbf{0.996} & \textbf{0.996} & \textbf{0.996} & {\ul 0.995} \\
SVHN         & 0.967          & \textbf{0.972} & {\ul 0.970}    & 0.968          & 0.965       \\
\midrule
CUB200       & \textbf{0.722} & 0.413          & 0.643          & {\ul 0.690}    & 0.672       \\
ISIC         & 0.684          & 0.585          & 0.680          & \textbf{0.695} & {\ul 0.694} \\
OxfordPets   & \textbf{0.891} & 0.540          & 0.801          & {\ul 0.850}    & 0.845       \\
StanfordCars & 0.696          & 0.424          & 0.678          & \textbf{0.718} & {\ul 0.703} \\
\bottomrule
\end{tabular}
\end{table}


A particularly notable observation is that the effect of adding intermediate distillation points is not linear with respect to the number of points used. 
On fine-grained datasets, the difference between \nameUnoB and the Blocks36 configuration, which only introduces a single additional distillation point, is considerably larger than the gap between the Blocks36 and Blocks246 configurations, despite the latter also differing by only one point. 
This suggests that intermediate knowledge provides substantial benefit even from a single additional distillation point, emphasizing the value of incorporating intermediate knowledge during distillation. 

A similar, though less pronounced, non-linear trend appears on classic datasets, where accuracy tends to decrease monotonically as more distillation points are added, from \nameUnoB down to \nameSeisB. 
On CIFAR100, however, this trend does not hold: both intermediate configurations outperform \nameUnoB and match the baseline, suggesting that the optimal number and placement of distillation points may also depend on dataset-specific characteristics rather than following a single universal pattern. 

Perhaps most notably, on most fine-grained datasets, the Blocks246 configuration matches or even outperforms \nameSeisB, despite distilling knowledge at only three blocks instead of all six. 
This indicates that not all blocks contribute equally useful information during distillation, and that distilling at every block may, in some cases, introduce noise or redundant supervision rather than additional benefit. 

\begin{figure*}[ht]
    \centering
    \includegraphics[width=\linewidth]{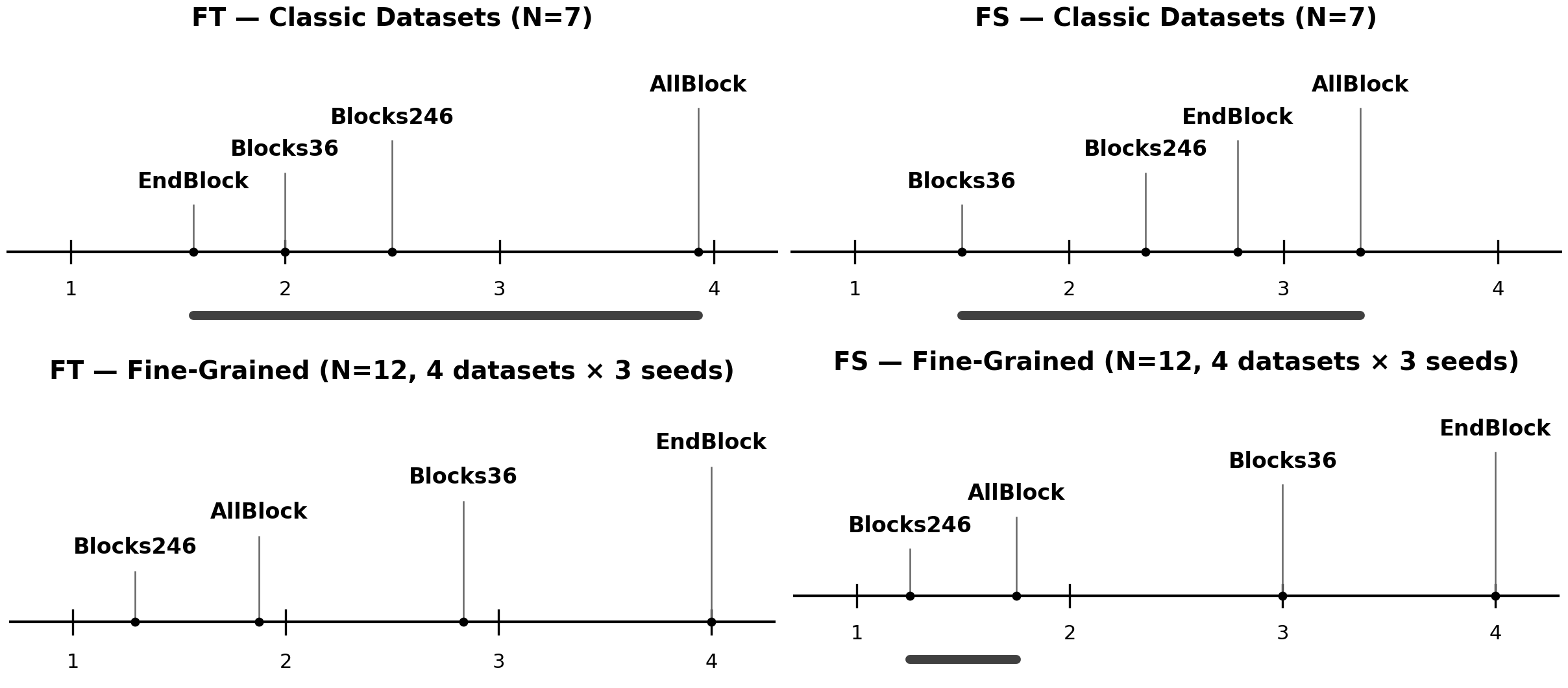}
    \caption{Critical difference diagrams comparing \nameUnoB, \nameSeisB, Blocks36 and Blocks246, for classic (top) and fine-grained (bottom) datasets, under FT (left) and FS (right). Lower ranks indicate better average accuracy; horizontal bars connect configurations with no statistically significant difference (Wilcoxon signed-rank test with Holm correction).}
    \label{fig:CD_Granularity}
\end{figure*}

These trends are further supported by a statistical analysis using Critical Difference (CD) diagrams, based on the Friedman test with Wilcoxon signed-rank post-hoc tests and Holm correction, comparing \nameUnoB, \nameSeisB, and both intermediate configurations under FT and FS ( Fig.~\ref{fig:CD_Granularity}). 
On classic datasets, although the Friedman test indicates a statistically significant ordering among configurations, the subsequent pairwise comparisons find no significant differences between any pair of configurations under either fine-tuning strategy. 
Under FT, the resulting ranking is monotonically related to the number of distillation points, from \nameUnoB to \nameSeisB, consistent with the trend already observed in \autoref{tbl:Granularity}; however, given the lack of statistically significant pairwise differences, this ordering should be regarded as merely suggestive. 
Overall, these results suggest that, when sufficient training data is available, the number and placement of distillation points have at most a limited effect on the resulting performance. 

On fine-grained datasets, in contrast, the same ranking is observed under both FT and FS: Blocks246 achieves the best average rank, followed by \nameSeisB, Blocks36, and \nameUnoB. 
Under FT, all pairwise differences are statistically significant, whereas under FS, no significant difference is found between Blocks246 and \nameSeisB, though both remain significantly better than Blocks36 and \nameUnoB. 
This consistent ranking indicates that, unlike in the classic setting, the selection of distillation points has a substantial and consistent impact on performance when training data is limited, further reinforcing the importance of intermediate knowledge in this regime. 
Notably, Blocks246 outranks \nameSeisB significantly under FT, suggesting that distilling every block may be excessive even under data scarcity. 
It should be noted that, for fine-grained datasets, the assumption of independence between instances required by these tests is not strictly satisfied, since three random seeds per dataset were used to reach the minimum sample size required by the Friedman test; these results should therefore be interpreted with appropriate caution. 

\subsection{Explainability}
\label{sec:ExplainabilityResults}

The results discussed in the preceding sections establish that intermediate block-wise distillation is beneficial, particularly under data scarcity, but leave open the question of why this is the case and how knowledge propagates through the student network. 
To address this, we turn to the explainability techniques introduced in \autoref{sec:XAI_method}, applying them at the output of every block regardless of whether it was used as a distillation point.

We organise this analysis in two stages. 
First, in \autoref{sec:Attention}, we compare teacher and student attention maps at each block through cosine similarity for
the configurations discussed in \autoref{sec:FineTuning} (\nameUnoB and \nameSeisB) under both fine-tuning strategies. 
Second, in \autoref{sec:cka-gradcam}, we use CKA and Grad-CAM to characterise which blocks contribute most to effective distillation and how this informs the design of new intermediate configurations. 

\subsubsection{Attention Map Analysis} 
\label{sec:Attention}

\begin{figure*}[ht]
    \centering
    \includegraphics[width=\linewidth]{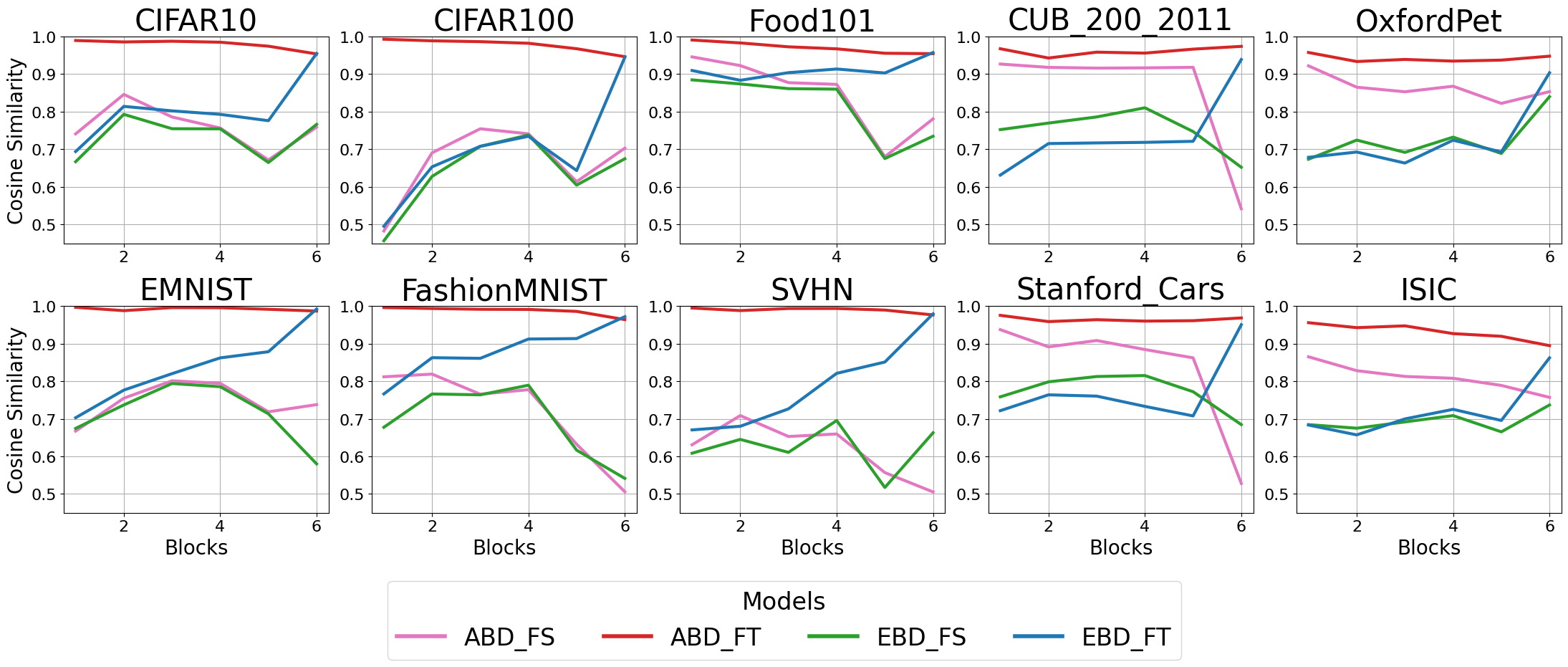}
    \caption{Block-wise cosine similarity between the representations of the fine-tuned teacher and the student, for the two extreme block-wise configurations (\nameUnoB, \nameSeisB) under both fine-tuning strategies (FT, FS), across all datasets.}
    \label{fig:Exp_cosine}
\end{figure*}

 Fig.~\ref{fig:Exp_cosine} shows the block-wise cosine similarity between teacher and student attention maps, computed for \nameUnoB and \nameSeisB under both fine-tuning strategies (MNIST is excluded, as it adds little additional insight). 
Similarity varies across blocks and configurations, with some models maintaining a consistently high similarity throughout the network and others fluctuating substantially, particularly in the earlier blocks. 
Within each configuration, the fine-tuning strategy governs the overall level and stability of that similarity, with FT models consistently achieving higher values than FS. 
\nameSeisB\_FT illustrates this most clearly, achieving near-perfect similarity with the teacher across all blocks and  datasets, including fine-grained ones. 
Since this configuration already provides explicit distillation at every block, this behaviour is expected.
In contrast, \nameUnoB models, under both fine-tuning strategies, display a markedly different pattern: similarity increases progressively across blocks, reaching its highest values only at the last one, where direct supervision is applied. 
At the final block, its similarity to the teacher reaches values comparable to \nameSeisB\_FT, even though \nameUnoB\_FT gets no KD supervision at any of the earlier blocks. 
However, the intermediate blocks remain noticeably less similar to the teacher than in \nameSeisB\_FT, confirming that KD at the intermediate blocks is needed to achieve that level of similarity.

Comparing \nameSeisB\_FS and \nameSeisB\_FT, while \nameSeisB\_FS also receives explicit supervision at every block, its similarity to the teacher is noticeably lower and less stable than that of \nameSeisB\_FT, and in several fine-grained datasets it drops sharply at the last block. 
A plausible explanation is that, in this case, 
training the classifier also unfreezes the rest of the student, partially undoing the block-wise alignment and allowing intermediate representations to drift without a fine-tuned teacher to anchor them.

Taken together, these results show that the two factors play complementary roles: the distillation configuration determines where along the network the student's representations converge towards the teacher's, while the fine-tuning strategy determines how strong and stable that convergence is. 
This distinction echoes the differences in accuracy observed in \autoref{sec:FineTuning}, where a fine-tuned teacher consistently yielded better results across configurations. 
However, attention maps alone offer only a coarse view of this alignment, motivating a closer examination through CKA and Grad-CAM in the following section.

\subsubsection{CKA and Grad-CAM Analysis} 
\label{sec:cka-gradcam}

We now turn to a more detailed analysis of how knowledge propagates across blocks, combining CKA and Grad-CAM. 
We first examine CKA, which allows us to compare full block-wise representations between teacher and student, before turning to Grad-CAM to assess how this translates into class-discriminative information. 

\begin{figure*}[t!]
    \centering
    \begin{subfigure}[b]{0.48\textwidth}
        \centering
        \includegraphics[width=\linewidth]{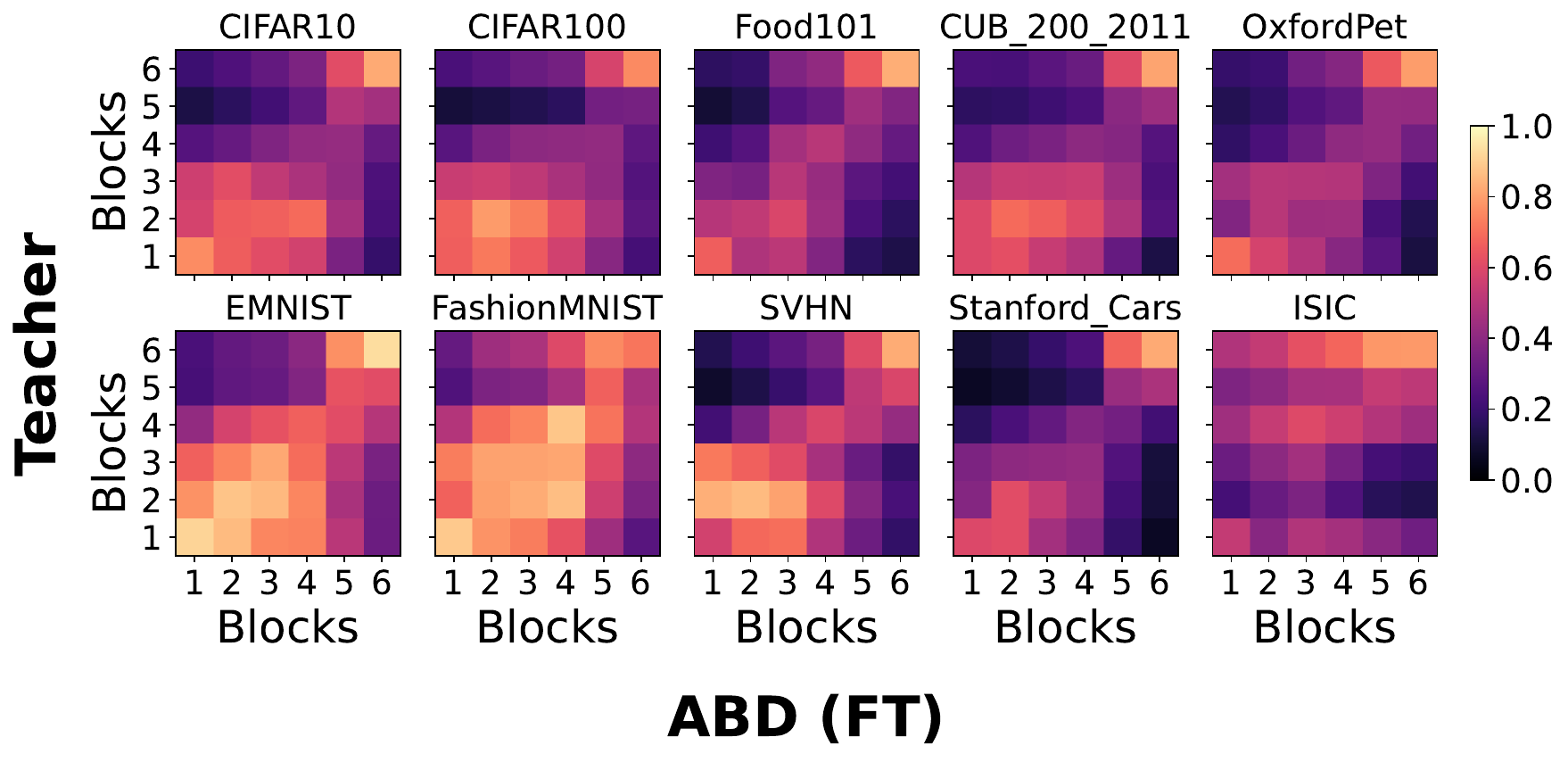}
        \caption{\fullnameSeisB (\nameSeisB)}
        \label{fig:CKA_AllBlock}
    \end{subfigure}%
    ~ 
    \begin{subfigure}[b]{0.48\textwidth}
        \centering
        \includegraphics[width=\linewidth]{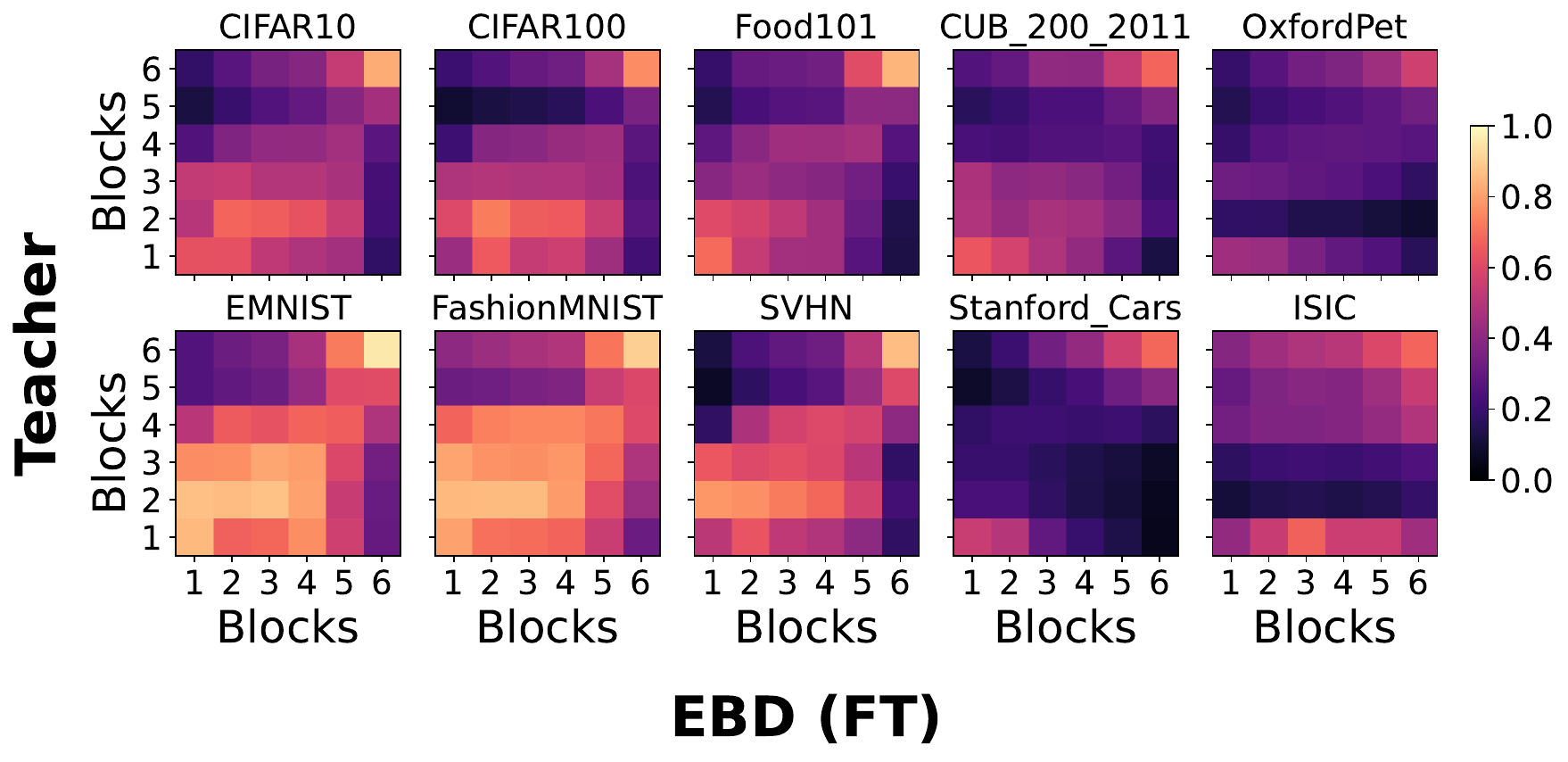}
        \caption{\fullnameUnoB (\nameUnoB)}
        \label{fig:CKA_EndBlock}
    \end{subfigure}
    \caption{Block-wise CKA between the representation of the fine-tuned teacher and the extreme KD configurations}
    \label{fig:CKA}
\end{figure*}

 Fig.~\ref{fig:CKA} shows the block-wise CKA between the fine-tuned teacher and the \nameSeisB\_FT and \nameUnoB\_FT, respectively, computed across all block pairs. 
On classic datasets, both configurations exhibit a broadly similar pattern, with relatively high CKA values concentrated around the diagonal, indicating that corresponding blocks share a comparable amount of representational structure regardless of whether intermediate supervision is applied. 
On fine-grained datasets, however, this similarity holds only for \nameSeisB\_FT: under \nameUnoB\_FT, CKA values remain low across almost the entire grid, including at the final block, directly KD-supervised block, consistent with the accuracy degradation reported in \autoref{sec:FineTuning}. 
Notably, this same block showed high attention-map similarity in \autoref{sec:Attention}, suggesting that activation-level alignment does not necessarily translate into structural alignment.

Having examined the representational structure recovered by the student through CKA, we now turn to Grad-CAM to assess whether a similar pattern holds for class-discriminative information. 
\autoref{tbl:GradCAM-CIFAR10} and \autoref{tbl:GradCAM-OxfordPet} show the block-wise Grad-CAM cosine similarity to the fine-tuned teacher and accuracy for \nameUnoB and \nameSeisB under both fine-tuning strategies, on a representative classic and fine-grained dataset, respectively (CIFAR10 and OxfordPet). 
The remaining datasets follow the same pattern within each group. 

On CIFAR10 (\autoref{tbl:GradCAM-CIFAR10}), \nameUnoB\_FT achieves Grad-CAM similarity comparable to, and at the final blocks even higher than, \nameSeisB\_FT, despite receiving no distillation signal at any of the earlier blocks. 
This reinforces the pattern already observed through CKA: when sufficient data is available, the student appears able to recover class-discriminative behaviour close to the teacher's largely on its own. 
On OxfordPet, in contrast, \nameSeisB\_FT clearly outperforms \nameUnoB\_FT across almost all blocks.
This mirrors the accuracy gap between the two students and reinforces the benefit of intermediate knowledge under data scarcity. 
Notably, even \nameSeisB\_FS, which distils from a teacher that has not been fine-tuned, still achieves higher similarity than \nameUnoB, suggesting that having intermediate distillation points matters more for recovering class-discriminative behaviour than whether the teacher itself has been fine-tuned. 

A particularly consistent pattern, observed across both  CKA (Fig.~\ref{fig:CKA_AllBlock}, Fig.~\ref{fig:CKA_EndBlock}) and Grad-CAM (\autoref{tbl:GradCAM-CIFAR10}, \autoref{tbl:GradCAM-OxfordPet}), is that accuracy and similarity at the last three blocks tend to move together. 
That is, models with higher accuracy also show higher similarity at these blocks, while the first three vary comparatively little regardless of overall performance. 

Taken together, the CKA and Grad-CAM analysis suggest that the last three blocks carry the information most directly responsible for classification performance, while the first three appear to encode more general, easily diluted knowledge that may not require individual distillation supervision. 

\begin{table}[ht]
\addtolength{\tabcolsep}{0.3em}
\centering
\caption{Average block-wise Grad-CAM cosine similarity to the fine-tuned teacher and average accuracy for the extreme block-wise configurations (\nameUnoB\_FT and \nameSeisB\_FT) on CIFAR10.}
\label{tbl:GradCAM-CIFAR10}
\addtolength{\tabcolsep}{3pt}
\begin{tabular}{crrrr}
\cmidrule{2-5}
         & \multicolumn{2}{c}{\nameUnoB}                   & \multicolumn{2}{c}{\nameSeisB}                   \\
\toprule
Blocks   & \multicolumn{1}{c}{FT} & \multicolumn{1}{c}{FS} & \multicolumn{1}{c}{FT} & \multicolumn{1}{c}{FS} \\
\midrule
1        & {\ul 0.388}            & 0.363                  & \textbf{0.449}         & 0.387                  \\
2        & {\ul 0.455}            & 0.444                  & \textbf{0.458}         & 0.443                  \\
3        & \textbf{0.456}         & {\ul 0.434}            & 0.403                  & 0.425                  \\
4        & {\ul 0.466}            & 0.423                  & \textbf{0.488}         & 0.338                  \\
5        & \textbf{0.495}         & {\ul 0.400}            & 0.291                  & 0.290                  \\
6        & \textbf{0.932}         & 0.753                  & {\ul 0.907}            & 0.690                  \\
\midrule
\midrule
Accuracy & \textbf{0.914}         & 0.873                  & {\ul 0.898}            & 0.889                 \\
\bottomrule
\end{tabular}
\end{table}

\begin{table}[ht]
\addtolength{\tabcolsep}{0.3em}
\centering
\caption{Average block-wise Grad-CAM cosine similarity to the fine-tuned teacher and average accuracy for the extreme block-wise configurations (\nameUnoB\_FT and \nameSeisB\_FT) on OxfordPet.}
\label{tbl:GradCAM-OxfordPet}
\addtolength{\tabcolsep}{3pt}
\begin{tabular}{crrrr}
\cmidrule{2-5}
                             & \multicolumn{2}{c}{\nameUnoB}                   & \multicolumn{2}{c}{\nameSeisB}                   \\
\toprule
\multicolumn{1}{c}{Blocks}   & \multicolumn{1}{c}{FT} & \multicolumn{1}{c}{FS} & \multicolumn{1}{c}{FT} & \multicolumn{1}{c}{FS} \\
\midrule
1                            & 0.386                  & 0.387                  & \textbf{0.427}         & {\ul 0.415}            \\
2                            & 0.402                  & {\ul 0.411}            & 0.376                  & \textbf{0.429}         \\
3                            & \textbf{0.396}         & {\ul 0.394}            & 0.361                  & 0.369                  \\
4                            & 0.378                  & 0.401                  & \textbf{0.434}         & {\ul 0.411}            \\
5                            & 0.375                  & 0.359                  & {\ul 0.439}            & \textbf{0.440}         \\
6                            & 0.747                  & 0.570                  & \textbf{0.923}         & {\ul 0.754}            \\
\midrule
\midrule
\multicolumn{1}{c}{Accuracy} & 0.540                  & 0.462                  & \textbf{0.845}         & {\ul 0.797}        \\
\bottomrule
\end{tabular}
\end{table}

\subsection{Explainability-Guided Student}
\label{sec:Blocks3456}

Motivated by the findings in \autoref{sec:ExplainabilityResults}, we introduce the student Blocks3456, in which the first three blocks are trained as a single group only through Block 3, while Blocks 4, 5, and 6 are distilled individually. 
We also evaluate its complementary counterpart, Blocks1236, which applies the same grouping logic in reverse, to isolate whether the benefit comes from specific blocs selected rather than merely their number. 

\autoref{tbl:4B_Results} shows the results for models Blocks3456 and Blocks1236 against the two extreme configurations. 
On classic datasets, Blocks1236 closely tracks \nameUnoB, consistent with the findings of \autoref{sec:ExplainabilityResults}: since early-block information dilutes regardless of supervision, densely distilling them adds little, making  Blocks1236 behave similarly to distilling only the last block. 
Blocks3456, in turn, matches \nameSeisB on these datasets, suggesting that distilling intermediate blocks may be detrimental, potentially over-constraining the student's representations during training. 
On fine-grained datasets, Blocks3456 achieves the best results, confirming that concentrating supervision on the last blocks is what matters. 
Blocks1236, in contrast, behaves similarly to Blocks36 (\autoref{tbl:Granularity}) and remains close to \nameUnoB: individually supervising the early blocks brings little extra benefit over simply grouping them. 

Notably, Blocks3456\_FS matches or slightly outperforms all other FS models on classic datasets, and clearly does so on fine-grained ones. 
This indicates that distilling a subset of blocks, rather than every block, can be sufficient for strong performance — a favourable time-accuracy trade-off particularly relevant since FS does not rely on a teacher already fine-tuned to the target dataset. 
On classic datasets, this small margin is consistent with the pattern already observed throughout this work: differences between distillation schemes are minor when sufficient training data is available.

\begin{table*}[ht]
\addtolength{\tabcolsep}{0.3em}
\centering
\caption{Average accuracy comparing \fullnameUnoB, Blocks1236, Blocks3456, and \fullnameSeisB}
\label{tbl:4B_Results}
\addtolength{\tabcolsep}{3pt}
\begin{tabular}{lrrrrrrrr}
\cmidrule{2-9}
             & \multicolumn{2}{c}{EBD}                         & \multicolumn{2}{c}{Blocks1236}                  & \multicolumn{2}{c}{Blocks3456}                  & \multicolumn{2}{c}{ABD}                         \\
\toprule
Datasets     & \multicolumn{1}{c}{FT} & \multicolumn{1}{c}{FS} & \multicolumn{1}{c}{FT} & \multicolumn{1}{c}{FS} & \multicolumn{1}{c}{FT} & \multicolumn{1}{c}{FS} & \multicolumn{1}{c}{FT} & \multicolumn{1}{c}{FS} \\
\midrule
CIFAR10      & \textbf{0.914}         & 0.873                  & {\ul 0.911}            & 0.885                  & 0.900                    & 0.891                  & 0.898                  & 0.889                  \\
CIFAR100     & 0.774                  & 0.733                  & \textbf{0.783}         & 0.740                  & {\ul 0.778}            & 0.763                  & 0.774                  & 0.743                  \\
EMNIST       & \textbf{0.907}         & 0.897                  & \textbf{0.907}         & 0.896                  & {\ul 0.905}            & 0.897                  & {\ul 0.905}            & 0.896                  \\
FashionMNIST & \textbf{0.944}         & 0.936                  & {\ul 0.943}            & 0.934                  & 0.937                  & 0.936                  & 0.937                  & 0.934                  \\
Food101      & \textbf{0.812}         & 0.751                  & {\ul 0.801}            & 0.750                  & 0.792                  & 0.753                  & 0.784                  & 0.734                  \\
MNIST        & \textbf{0.996}         & {\ul 0.995}            & {\ul 0.995}            & 0.994                  & {\ul 0.995}            & {\ul 0.995}            & {\ul 0.995}            & {\ul 0.995}            \\
SVHN         & \textbf{0.972}         & 0.957                  & {\ul 0.970}             & 0.957                  & 0.966                  & 0.956                  & 0.965                  & 0.955                  \\
\midrule
CUB200       & 0.413                  & 0.364                  & 0.620                   & 0.608                  & \textbf{0.682}         & 0.664                  & {\ul 0.672}            & 0.659                  \\
ISIC         & 0.585                  & 0.488                  & 0.69                   & 0.646                  & 0.684                  & \textbf{0.705}         & {\ul 0.694}            & 0.687                  \\
OxfordPets   & 0.540                  & 0.462                  & 0.79                   & 0.736                  & \textbf{0.854}         & 0.816                  & {\ul 0.845}            & 0.797                  \\
StanfordCars & 0.424                  & 0.449                  & 0.657                  & 0.697                  & 0.721                  & \textbf{0.748}         & 0.703                  & {\ul 0.744}           \\
\bottomrule
\end{tabular}
\end{table*}


\section{Conclusions}
\label{sec:Conclusions}

This work studied block-wise knowledge distillation between a homogeneous student and an EfficientNet-B0 teacher across 
both classic and fine-grained, data-scarce datasets, focusing on how the number and placement of intermediate distillation points interact with data availability and dataset granularity. 
Our results first establish a practical fine-tuning strategy: fine-tuning the teacher alone (FT) generally achieves the best results, while fine-tuning the student alone (FS) offers a competitive, lower-cost alternative, with jointly fine-tuning both bringing no additional benefit despite its higher cost (RQ1). 
The teacher's pre-training domain also matters: under data scarcity, a teacher whose pre-training domain differs substantially from the target task may fail to adapt sufficiently, limiting the quality of knowledge it can transfer. 
On classic, data-abundant datasets, no configuration offered a consistent advantage over \nameUnoB, which remains a sound default given its lower computational cost (RQ2). 
Under data scarcity, however, intermediate supervision 
narrows the gap, though non-linearly and not always enough to surpass the teacher:  
a single additional point already recovers most of it
, with diminishing and occasionally negative returns beyond that, as densely distilling every block risks over-constraining the student rather than aiding it (RQ2, RQ4). 
Our DA and data reduction experiments further suggest that this benefit is driven primarily by the number of available instances per class, rather than by complexity itself: augmenting fine-grained datasets narrows the gap between \nameUnoB and \nameSeisB, while reducing the training data on classic datasets reproduces the same pattern observed on fine-grained ones (RQ3). 

Our explainability analysis partly explains this pattern: 
early blocks encode more general representations weakly tied to classification performance, while the last blocks carry most of the class-discriminative information (RQ5). 
Students concentrating supervision on these later blocks, such as Blocks3456, matched or exceeded \nameSeisB while using fewer distillation points, and proved especially effective in the FS setting — closer to real-world scenarios lacking a fine-tuned teacher — outperforming all other FS models on both classic and fine-grained datasets (RQ6).

\begin{figure}
    \centering
    \includegraphics[width=0.8\linewidth]{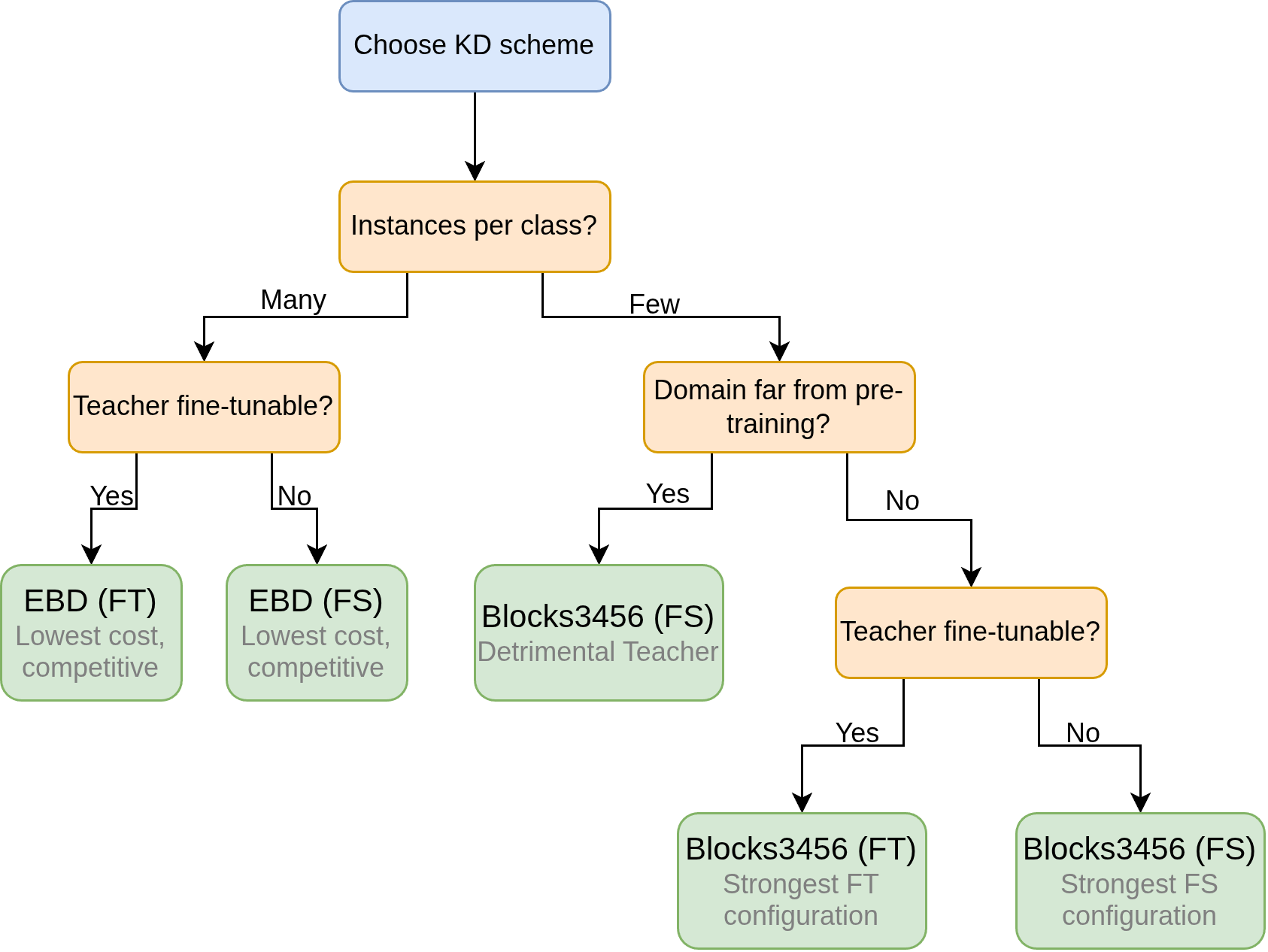}
    \caption{Practical guideline for selecting a distillation scheme.}
    \label{fig:Decision}

\end{figure}

Building on these findings, we distil them into a practical guideline for selecting a distillation scheme ( Fig.~\ref{fig:Decision}) depending on the scenario.
As applications increasingly demand compact models under limited data, guiding distillation by where knowledge is transferred, rather than only how much, offers a lightweight and broadly applicable path towards more data-efficient students.
Although output-level distilling remains competitive, our results demonstrate that intermediate-layer distillation provides a substantial improvement in the low-data regimes characteristic of most real-world applications, making it an effective choice for data-constrained settings.

\bibliographystyle{unsrt}
\bibliography{biblio}  






\end{document}